\documentclass[10pt,twocolumn,letterpaper]{article}

\usepackage{iccv}      
\usepackage{algorithmic}
\usepackage{amsmath}
\usepackage{colortbl} 
\usepackage{xcolor}   
\usepackage{multirow}
\usepackage{xcolor}
\usepackage{enumitem}

\usepackage[linesnumbered,ruled,vlined, noend]{algorithm2e}

\SetCommentSty{mycommfont}

\usepackage{caption}
\usepackage{xspace}

\definecolor{iccvblue}{rgb}{0.21,0.49,0.74}
\usepackage[pagebackref,breaklinks,colorlinks,allcolors=iccvblue]{hyperref}
\usepackage{hyperref}

\title{Q\&C: When Quantization Meets Cache in Efficient Image Generation}

\author{\textbf{Xin Ding}\textsuperscript{1} \quad \textbf{Xin Li}\textsuperscript{1} \quad \textbf{Haotong Qin}\textsuperscript{2} \quad \textbf{Zhibo Chen}\textsuperscript{1}\\
\textsuperscript{1}University of Science and Technology of China \quad \textsuperscript{2}ETH Zürich, Switzerland \\
{\tt\small xinding64@mail.ustc.edu.cn,  haotong.qin@pbl.ee.ethz.ch, \{xin.li, chenzhibo\}@ustc.edu.cn}
}

\begin{document}
\maketitle

\begin{abstract}
Quantization and cache mechanisms are typically applied individually for efficient Diffusion Transformers (DiTs), each demonstrating notable potential for acceleration. However, the promoting effect of combining the two mechanisms on efficient generation remains under-explored. Through empirical investigation, we find that the combination of quantization and cache mechanisms for DiT is not straightforward, and two key challenges lead to severe catastrophic performance degradation:
(i) the sample efficacy of calibration datasets in post-training quantization (PTQ) is significantly eliminated by cache operation; (ii) the combination of the above mechanisms introduces more severe exposure bias within sampling distribution, resulting in amplified error accumulation in the image generation process. In this work, we take advantage of these two acceleration mechanisms and propose a hybrid acceleration method by tackling the above challenges, aiming to further improve the efficiency of DiTs while maintaining excellent generation capability.
Concretely, a temporal-aware parallel clustering (TAP) is designed to dynamically improve the sample selection efficacy for the calibration within PTQ for different diffusion steps. A variance compensation (VC) strategy is derived to correct the sampling distribution. It mitigates exposure bias through an adaptive correction factor generation. Extensive experiments have shown that our method has accelerated DiTs by 12.7 $\times$ while preserving competitive generation capability. The code will be available at \href{https://github.com/xinding-sys/Quant-Cache}{https://github.com/xinding-sys/Quant-Cache}.

\end{abstract}

\section{Introduction}
The rapid rise of Diffusion Transformers (DiTs) \cite{dit} has driven significant breakthroughs in generative tasks, particularly in image generation \cite{croitoru2023diffusion,yang2023diffusion}. With their transformer-based architecture \cite{carion2020end,touvron2021training,xie2021segformer}, DiTs offer superior scalability and performance \cite{videoworldsimulators2024}. However, their widespread adoption is hindered by the immense computational complexity and large parameter counts.  For instance, generating a 512$\times$512 resolution image using DiTs can take more than 20 seconds and 105 Gflops on an NVIDIA RTX A6000 GPU \cite{wu2024ptq4dit}. This substantial requirement makes them unacceptable or impractical for real-time applications, especially as model sizes and resolutions continue to increase \cite{liu2024hq,zhao2024vidit}.
\begin{figure}
    \centering
    \includegraphics[width=0.9\linewidth]
    {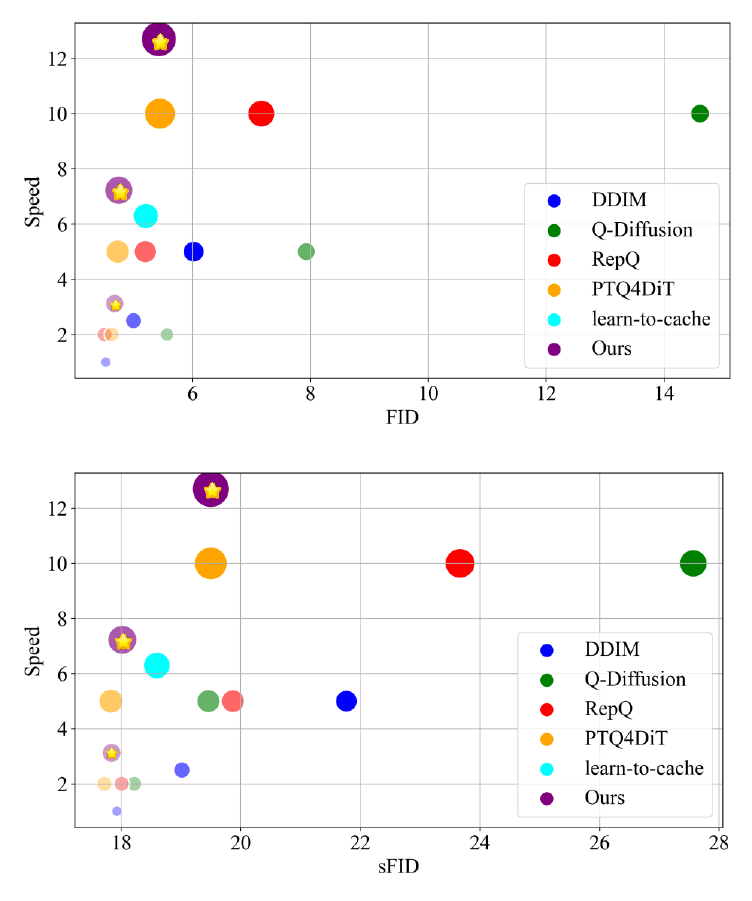}
    \caption{Efficiency-versus-efficacy trade-off across different settings. Bubble size represents the ratio of relative speed-up to generative quality compared to the DDPM baseline at 250 timesteps. We compare various methods in terms of FID (top) and sFID (bottom) performance across 50, 100, and 250 timesteps. Our method consistently appears in the upper-left region across all settings, achieving maximum acceleration while preserving generative quality.}
    \label{fig:comparation}
\end{figure}

Quantization \cite{nagel2021white,nagel2020up,liu2023oscillation} and cache \cite{xu2018deepcache,wimbauer2024cache,so2023frdiff}, as two acceleration mechanisms, have been initially explored to alleviate the computational burden of DiTs \cite{wu2024ptq4dit,ma2024learning,selvaraju2024fora,chen2024delta} individually. Quantization accelerates models by converting weights and activations into lower-bit formats, significantly reducing inference time and memory usage. In particular, post-training quantization (PTQ) \cite{he2023efficientdm}, as a quantization paradigm, merely requires a small calibration dataset to eliminate quantization errors, which is effective but resource-friendly for DiTs compared with quantization-aware training (QAT) \cite{lu2024terdit}. In contrast, the cache mechanism intends to exploit the reusability of history features during the diffusion process to obviate the computational costs for inference, thereby being another popular way to accelerate the DiTs. The commonly used strategies for cache exploit the repetitive nature of the diffusion process, storing and reusing the feature from layers like attention and MLP across different denoising steps. 
\begin{figure*}
    \centering
    \includegraphics[width=1.\linewidth]
    {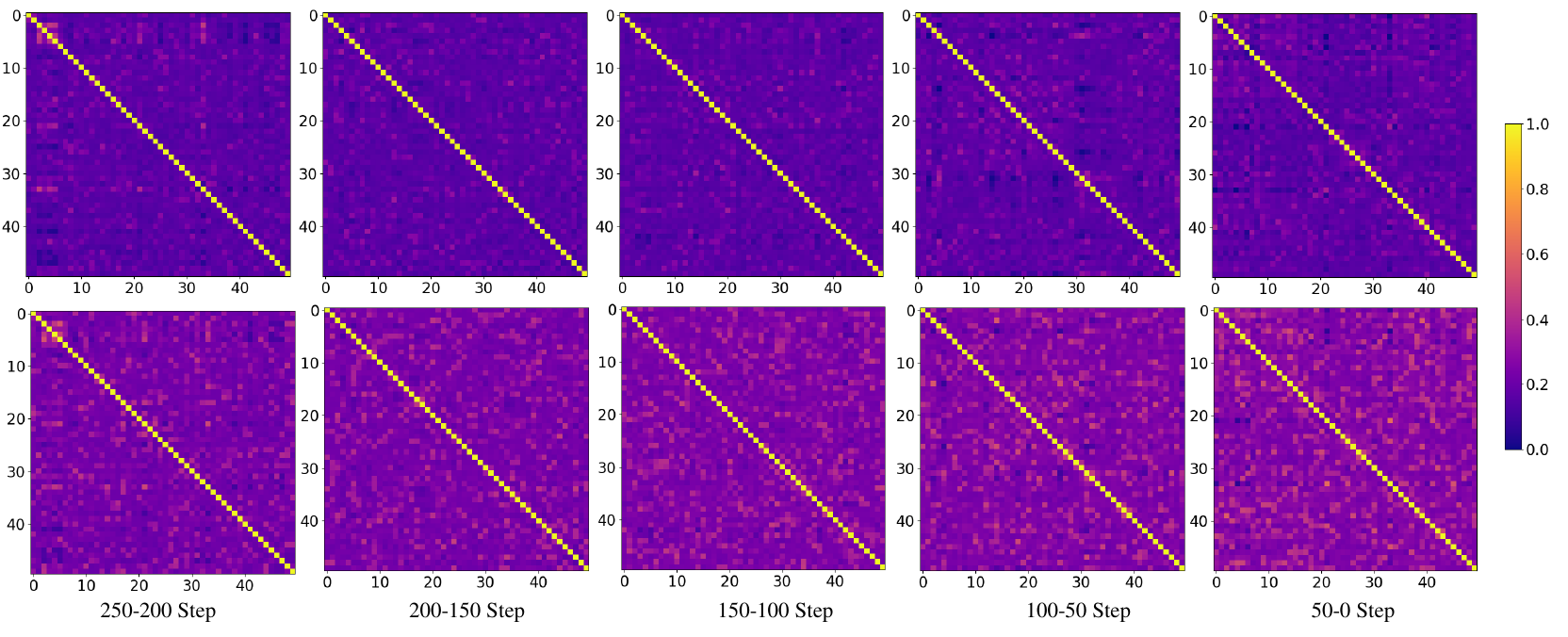}
    \caption{Cosine similarity analysis across time steps in DiT for calibration data. This visualization is based on a 250-step DDIM sampling process. Calibration data were collected both without (up) and with (bottom) cache; samples positioned further to the right represent data closer to the final step $x_0$. The heatmap reveals high similarity in calibration datasets when quantization meets cache, particularly in later diffusion stages. This observation motivates our calibration strategy, highlighting a clear requirement to reduce redundancy and improve efficacy.}
    \label{fig:similarity}
\end{figure*}

Despite the effectiveness of both quantization and cache mechanisms, it remains under-explored ``whether integrating these two mechanisms can further boost the efficiency for DiTs?"
However, when quantization meets cache, a notable decline in the generative quality of DiTs is observed even though this achieves impressive acceleration benefits. To address this question, we conduct an in-depth analysis of the sampling process in DiTs and identify two crucial factors contributing to the decline in performance. Firstly, as shown in Figure~\ref{fig:similarity}, we surprisingly find that the sample similarity in the calibration dataset used for PTQ is dramatically increased by cache operation, leading to a marked reduction in sample efficacy. Moreover, this reduction becomes progressively more severe with the increase of diffusion steps, which compromises the effectiveness of PTQ due to the insufficient coverage of the overall generative distribution. Secondly, the synergy of quantization and cache results in more severe exposure bias (See supplementary materials for a detailed definition) within the sampling distribution, a problem that is less pronounced when exploring either quantization or cache individually, which can be observed in Fig.~\ref{fig:exposure bias}. Besides, the exposure bias leads to an accumulated shift in the distribution variance of denoised output when the sampling iterations increase in DiTs.  

To restore the generation capability of DiTs while keeping enhanced acceleration achieved through combing quantization and cache mechanisms, we tackle the above challenges by developing two essential techniques, constituting our hybrid acceleration mechanism: (i) temporal-aware parallel clustering (TAP) and (ii) distribution variance compensation (VC).

In particular, our TAP aims to restore the reduced sample efficacy in calibration datasets caused by cache operation, thereby 
enabling more accurate identification and correction of quantization errors. 
Notably, a na\"ive method to overcome the reduction of sample efficacy is to increase the dataset size. However, it will introduce excessive redundant data and unnecessary computational costs. In contrast, our TAP constructs the calibration dataset by dynamically selecting the most informative and distinguishable samples from large-scale datasets in an efficient clustering manner. Unlike traditional spectral clustering, which  suffers from unaffordable computational complexity at $O(n^3)$ \cite{yan2009fast,li2011time,chen2011large} and even $O(n^2)$ with accelerated/optimized algorithms \cite{halko2011finding,feng2018faster,martin2018fast}.
TAP integrates temporal sequences with data distribution to enable parallel processing subsample of size $r$ and reduce computational costs. This design leverages the time-sensitive nature of diffusion calibration datasets, as highlighted in recent studies \cite{li2023q,li2024q}, allowing for effective clustering and sampling that better represents the overall distribution without excessive redundancy with a computational complexity of $O(rn)$, where $r\ll n$.

Our in-depth analysis of the image generation process reveals a strong link between image variance and exposure bias, as shown in Sec.\ref{sec:challenge2}. To address this, we propose the VC, a tailored approach that adaptively mitigates exposure bias through variance correction. Unlike methods that introduce an additional neural network to predict errors in corrupted estimations \cite{wimbauer2024cache}, our approach requires no additional training. Instead, it utilizes a small batch of intermediate samples to compute a reconstruction factor, which adaptively corrects feature variance at each timestep. This method effectively reduces exposure bias, resulting in notable improvements in overall model performance.

The contributions of this paper can be summarized as follows:
\begin{itemize}
    \item We are the first to investigate the combined use of quantization and caching techniques in DiTs, demonstrating the substantial potential of this approach to alleviate computational burdens.
    \item We identify two critical challenges when integrating quantization and cache: (1) the generation of highly redundant samples in calibration datasets; and (2) the emergence of exposure bias caused by distributional variance shifts in the model's output, which becomes exacerbated over iterations.
    \item We propose two novel methods: (1) TAP dynamically selects informative and distinct samples from large-scale datasets to optimize calibration dataset efficacy. (2) VC, an adaptive approach that mitigates exposure bias by correcting feature variance at each timestep requiring no additional training.
    \item Extensive empirical results demonstrate that our approach accelerates diffusion image generation by up to 12.7$\times$ while maintaining comparable generative quality.    
    \end{itemize}

\section{Background and Motivation}
\subsection{Quantization and Cache}
Quantization, a pivotal stage in model deployment, has often been scrutinized for its ability to reduce memory footprints and inference latencies.  Typically, its quantizer $Q(X|b)$ is defined as follows:
\begin{equation}
    \begin{aligned}
        Q(X|b)=\mathrm{clip}(\left \lfloor \frac{X}{s}  \right \rceil+z,0,2^b-1  )
    \end{aligned}
\end{equation}
Where $s$ (scale) and $z$ (zero-point) are quantization parameters determined by the lower bound $l$ and the upper bound $u$ of $X$,which are usually defined as follow:
\begin{equation}
    \begin{aligned}
        l = \mathrm{min}(X),u = \mathrm{max}(X)
    \end{aligned}
\end{equation}
\begin{equation}
    \begin{aligned}
        s = \frac{u-l}{2^b-1},z = \mathrm{clip}(\left \lfloor -\frac{l}{s}  \right \rceil+z,0,2^b-1  ) 
    \end{aligned}
\end{equation}
Using a calibration dataset and equations (2) and (3), we can derive the statistical information for $s$ and $z$. Previous research \cite{williams2024impact,lee2023enhancing,wu2023zeroquant,jaiswal2023compressing} has examined the performance of downstream tasks across a variety of models, compression methods, and calibration data sources. Their findings indicate that the choice of calibration data can significantly impact the performance of compressed models. 

Cache, a technique that leverages the repetitive nature of denoising steps in diffusion models, significantly reduces computational costs while maintaining the quality of generated samples. cache mechanisms operate by storing and reusing intermediate outputs during the sampling process, avoiding the need for redundant calculations at each step. The key parameter in this approach is the cache interval $N$, which dictates how often features are recomputed and cached. Initially, features for all layers are cached, and at each time step $t$, if $mod N = 0$, the model recomputes and updates the cache. For the following $N-1$ steps, the model reuses these cached features, bypassing the need for repeated full forward passes. This process efficiently reduces computational overhead, particularly in diffusion models, without sacrificing generative quality.

\subsection{Challenges in the Synergy of Quantization and cache in Efficient Image Generation}
\label{sect:challenge}
The remarkable performance of quantization and cache naturally leads us to consider the significant potential of their combination for enhancing the efficiency of DiTs. To this end, we conducted an in-depth analysis and identified two critical issues.

\paragraph{Challenge 1: Degradation in Calibration Dataset Effectiveness}
In diffusion quantization, previous works \cite{liu2024enhanced,zhao2024mixdq,li2023q} often randomly sample intermediate inputs uniformly across all time steps to generate a small calibration set. This strategy leverages the smooth transition between consecutive time steps, ensuring that a limited calibration set can still represent the overall distribution effectively \cite{li2024q}. \textbf{However, when quantization meets cache, this balance is disrupted, significantly reducing the effectiveness of the calibration dataset.} 

To visualize this issue, we followed the setup in \cite{wu2024ptq4dit} and constructed multiple calibration datasets, each consisting of 250-step samples. We then computed the cosine similarity between these samples and observed a substantial rise in similarity compared to non-cached scenarios (see Figure \ref{fig:similarity}). Furthermore, as the diffusion process approaches the final step $x_0$, sample similarity increases dramatically, with some exceeding 60\%. Paradoxically, these later-stage samples are more reliable and valuable for accurate calibration. \textbf{This indicates that a large portion of calibration samples, despite their computational cost, do not contribute additional useful information for quantization}, significantly reducing the overall effectiveness of the calibration dataset.

\paragraph{Challenge 2: Amplification of Exposure Bias}
\label{sec:challenge2}
\begin{figure}
    \centering
    \includegraphics[width=1.\linewidth]
    {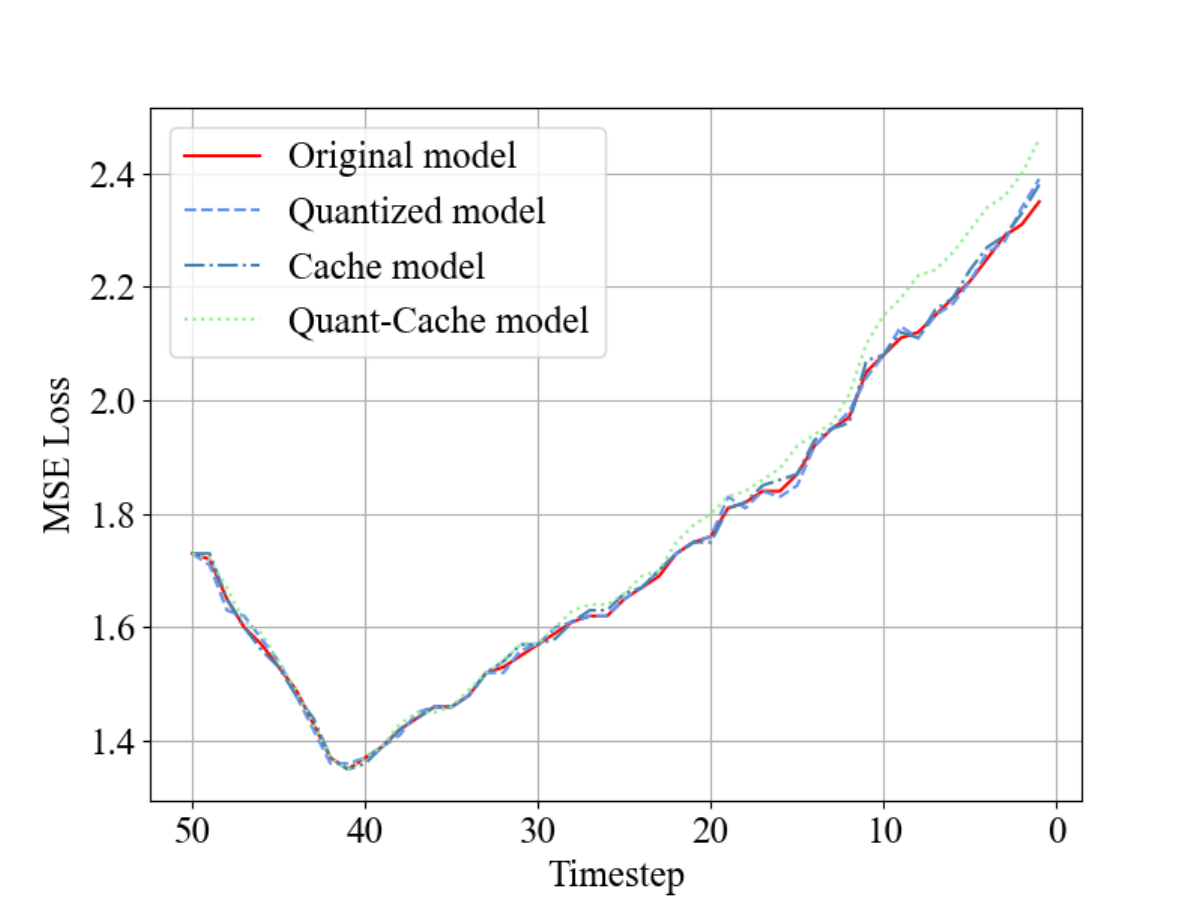}
    \caption{Analysis of exposure bias in DiT models. The mean squared errors between predicted samples and ground truth samples are computed at each time step. While the exposure bias remains relatively stable in both the cached and quantized models compared to the 50-timestep DiT, a noticeable increase in exposure bias is observed when quantization meets cache, leading to accumulation during the generation process.}
    \label{fig:exposure bias}
\end{figure}
Past research has consistently shown that exposure bias, resulting from the training-inference discrepancy, has a profound impact on text and image generation models \cite{ranzato2015sequence,schmidt2019generalization,rennie2017self,ning2023input}. Due to the presence of exposure bias, it gradually intensifies with the increase in the number of inference sampling steps, becoming a major cause of error accumulation \cite{li2023alleviating,li2023error} (See supplementary materials for a more detailed definition). To explore this further, we compared the changes in exposure bias under different acceleration methods and were surprised to find that \textbf{when quantization meets cache, exposure bias significantly worsens, whereas it does not occur when either quantization or cache is used in isolation, as shown in Figure \ref{fig:exposure bias}}.

\begin{figure}
    \centering
    \includegraphics[width=0.85\linewidth]
    {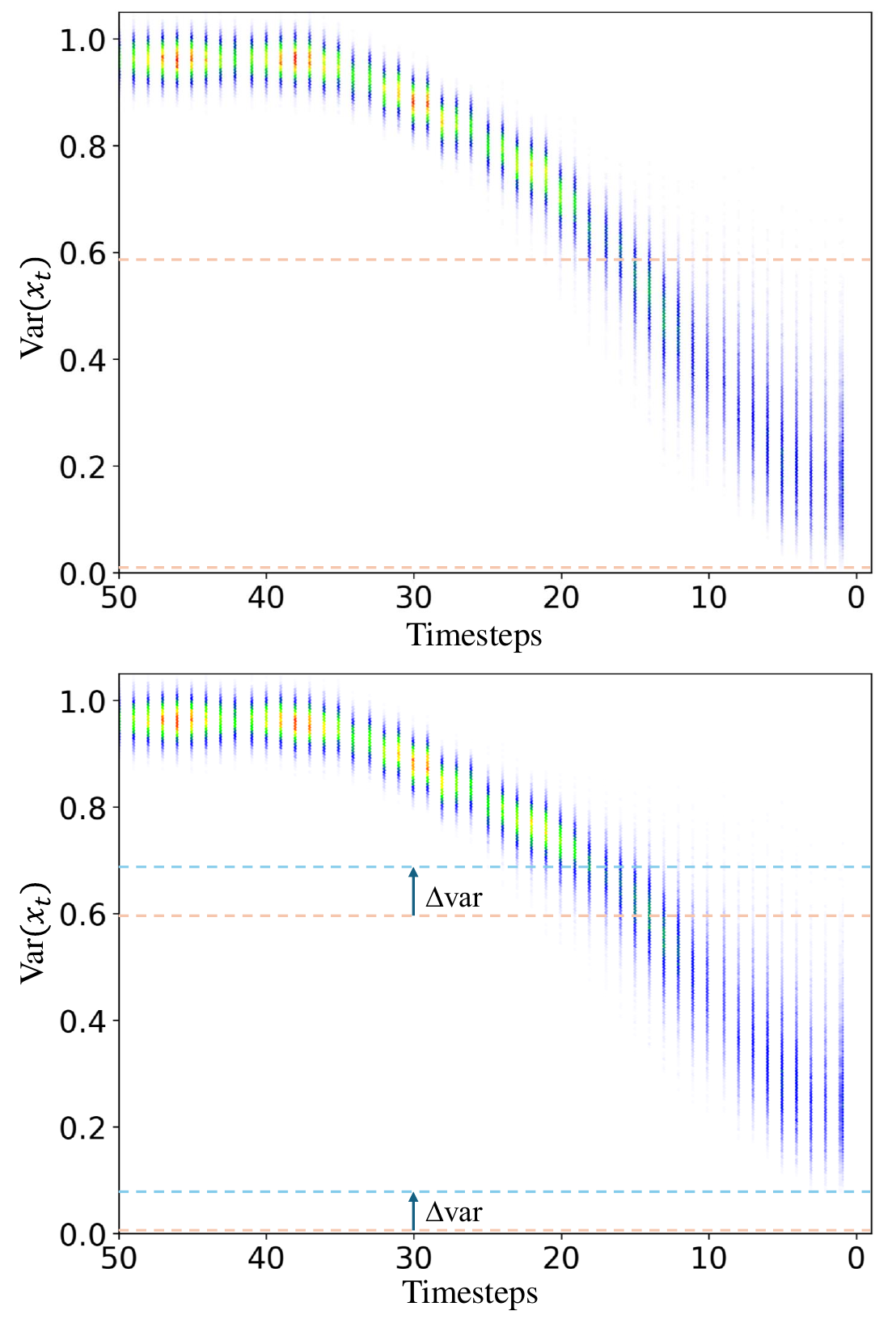}
    \caption{Comparision of the density distribution of the variance of 5000 samples from Imagenet across difference time steps. They illustrate the change in sample distribution variance at various time steps, shown for case without (top) and with (bottom) quant-cache. As the diffusion progresses, the variance of sample distribution starts to deviate towards Gassian white noise.}
    \label{fig:variance}
\end{figure}

To analyze the underlying causes, we examined the distributional changes over the generation process using 5,000 images. We observed that this Amplification is due to a change in variance. Specifically, as shown in Figure \ref{fig:variance}, at the beginning of the denoising process, the span of variance is narrow, and the changes in variance remain stable, fluctuating around 1. As the noise is gradually removed from the white noise, the variance distribution of the ground truth samples spans approximately (0, 0.6), reflecting the diversity of the sample distributions. However, \textbf{when considering the synergy of quantization and cache, the distribution shifts to the range of (0.1, 0.7), which aligns closely with the shift trend of exposure bias in Fig.\ref{fig:exposure bias}.} \textbf{We conducted the same experiment for the mean, but no similar phenomenon was observed, detailed analysis can be found in the supplementary materials.} This highlights the need to correct variance during the later stages of generation to mitigate its negative impact on exposure bias.

\section{Method}

\subsection{Temporal-Aware Parallel Clustering for Calibration}
In this section, we present Temporal-Aware Parallel Clustering (TAP), a novel method that integrates both spatial data distribution and temporal dynamics to address clustering challenges in datasets with complex feature interactions and inherent temporal patterns. TAP leverages parallel subsampling to efficiently combine spatial and temporal similarities, providing a robust approach for generating calibrated datasets.

\paragraph{Algorithm Overview} Given a dataset $T$ with $N$ samples, TAP reduces computational complexity through subsampling, followed by parallel processing across multiple subsampled sets. Each subsample is generated via random sampling, where the probability of selecting a sample is $ p_i = \frac{n}{N}$, with $n$ being the number of samples per subsample. By repeating this process, we obtain $m$ subsampled sets $\{S_1, S_2, \dots, S_m\}$. The parallel subsampling approach offers two key advantages: (1) it mitigates potential random noise and distributional biases within the dataset, and (2) it significantly improves computational efficiency.

For each subsampled set, a similarity matrix $A_{\text{final}}^{(i)}$ is constructed. Next, spectral clustering is applied to each weighted similarity matrix $A_{\text{final}}$ to detect communities.First, we compute the normalized Laplacian matrix for each parallel subsampling set,as follow:
\begin{equation}
    L^{(i)} = (D^{(i)}_r)^{\frac{1}{2}}A^{(i)}_{\text{final}}(D^{(i)}_c)^{\frac{1}{2}}\in\mathbb{R}^{N\times n} 
\end{equation}
Given the subsampled similarity matrix $A^{(i)}_{\text{final}}$, where $D$ is a diagonal matrix with the $i$-th diagonal element being $\sum_h A_{kh}$, for $1\leq k \leq N$, the degree matrices for the subsampled node set $S_i$ are defined as:
\begin{equation}
    D^{(i)}_r = \text{diag}\left \{ (D^{(i)}_{r,k})^N_{k=1} \right \} ,
    D^{(i)}_c = \text{diag}\left \{ (D^{(i)}_{c,h})^N_{h=1} \right \} 
\end{equation}
The top $k$ eigenvectors of $L$ are then extracted, and k-means clustering is performed on the rows of the resulting eigenvector matrix to produce the final clustering results. 

As the entire dataset is divided into $k$ categories, we can uniformly sample from these categories to construct the final calibration dataset, ensuring that its data distribution perfectly covers the overall distribution of the original dataset. The detailed algorithm flow is shown in Algorithm\ref{alg:TAP}.

\paragraph{Definition of Similarity Matrices $A_{\text{final}}^{(i)}$}
Drawing from the prior work \cite{li2023q,he2024ptqd}, datasets $T$ exhibit complex feature distributions and inherent temporal patterns. To account for these aspects, we construct a comprehensive similarity measure by combining spatial and temporal similarities. Specifically, for each subset $S_i$, we compute the data similarity matrix $A_{\text{data}}^{(i)}$ based on the feature space, and the temporal similarity matrix $A_{\text{time}}^{(i)}$, which captures temporal correlations. We then construct a weighted similarity matrix for each subsample, which combines both spatial and temporal similarities:
\begin{equation}
    A_{\text{final}}^{(i)} = \alpha A_{\text{spatial}}^{(i)} + (1-\alpha) A_{\text{temporal}}^{(i)}
\label{equ:similarity matrices}
\end{equation}
where $\alpha$ represents adjustable weights that balance the influence of data spatial and temporal properties.

Spatial Similarity Matrix $A^{(i)}_{\text{spatial}}$ captures the similarity between samples in terms of their data features. For each pair of samples $x_k$ and $x_h$ from the subsampled set $S_i$, the element $A^{(i)}_{\text{spatial},kh}$ represents how similar these two samples are based on their feature vectors, which could be defined as:
\begin{equation}
    A^{(i)}_{\text{data},kh} = \frac{x_k \cdot x_h}{\| x_k \| \| x_h \|}
\end{equation}

Temporal Similarity Matrix $A^{(i)}_{\text{temporal}}$ captures the similarity between samples based on their temporal relationships. For each pair of samples with timestamps $t_k$ and $t_h$, the element $A^{(i)}_{\text{temporal},kh}$ could be defined as:
\begin{equation}
    A^{(i)}_{\text{time},kh} = \exp\left(-|t_k - t_h|\right) 
\end{equation}

\vspace{-5mm}
\begin{algorithm}[ht]
\caption{Temporal-Aware Parallel Clustering (TAP)}
\label{alg:TAP}
\KwIn{Dataset $T$ with $N$ samples, samples per subsample $n$}
\KwOut{Cluster assignments for dataset $T$}

\For{$i = 1$ to $m$ \textbf{in parallel}}{
    Generate subsample $S_i$ from $T$ with $|S_i| = n$;
    
    Compute the spatial matrix $A^{(i)}_{\text{spatial}}$ for $S_i$:\;
    \Indp
    \For{each pair $(x_k, x_h) \in S_i$}{
        $A^{(i)}_{\text{spatial},kh} \gets \frac{x_k \cdot x_h}{\|x_k\| \|x_h\|}$\;
    }
    \Indm
    
    Compute the temporal matrix $A^{(i)}_{\text{temporal}}$ for $S_i$:\;
    \Indp
    \For{each pair with timestamps $(t_k, t_h) \in S_i$}{
        $A^{(i)}_{\text{temporal},kh} \gets \exp\left(-|t_k - t_h|\right)$\;
    }
    \Indm
    
    Combine spatiotemporal similarities $A^{(i)}_{\text{final}}$;
    
    Compute the degree matrices $D^{(i)}_r$ and $D^{(i)}_c$;
    
    Compute the normalized Laplacian matrix:\;
    $L^{(i)} \gets (D^{(i)}_r)^{\frac{1}{2}} A^{(i)}_{\text{final}} (D^{(i)}_c)^{\frac{1}{2}}$\;
    
    Extract the top $k$ eigenvectors of $L^{(i)}$ and perform k-means clustering on the eigenvector matrix rows\;
}

Aggregate cluster assignments from all subsamples to produce final clustering results\;
\end{algorithm}

\subsection{Variance Align for Exposure Bias }

Assume that a random variable $f$ follows a normal distribution, denoted as $f \sim \mathcal{N}(\mu, \sigma^2)$, where $\mu$ represents the mean and $\sigma^2$ denotes the variance. To alter the variance of $f$, we can apply a scaling transformation. If the objective is to modify the variance to a new value $\sigma_{\text{new}}^2$, the transformation can be defined as follows:
\begin{equation}
Y = \mu + \frac{\sigma_{\text{new}}}{\sigma} (f - \mu)
\end{equation}

In this formulation, $Y$  will conform to a new normal distribution given by $Y \sim \mathcal{N}(\mu, \sigma_{\text{new}}^2)$, where $\frac{\sigma_{\text{new}}}{\sigma}$ serves as the scaling factor for the variance adjustment. However, in practice, directly determining $\frac{\sigma_{\text{new}}}{\sigma}$ may not be feasible. Consequently, we introduce a timestep-dependent reconstruction scaling factor $\mathbf{K} \in \mathbb{R}^{St \times C}$ within the Intermediate samples $\hat{x}$, where $St$ indicates the number of denoising steps and $C$ signifies the number of channels corresponding to the estimated noise. The reconstructed Intermediate samples $\Tilde{x}_t$ at timestep $t$ can thus be represented as follows:
\begin{equation}
    \Tilde{x}_t = \mu_t + \mathbf{K}_t \cdot (\hat{x}_t-\mu_t)
\end{equation}
Where $(\cdot)$ represents the channel-wise multiplication. Next, we need to select a suitable optimization objective $\mathcal{L}$ to efficiently reconstruct feature $\Tilde{x}_t$.  

Mean Squared Error (MSE) is frequently employed to measure the discrepancy between the reconstructed feature $\tilde x_t$ and the target feature $x'_t$. However, MSE primarily assesses global numerical deviation and overlooks the channel-specific noise impact\cite{finkelstein2019fighting,nagel2019data}. To capture these nuances, we enhance the MSE criterion with the inverse root quantization-to-noise ratio (rQNSR)\cite{finkelstein2019fighting}, the optimization objective can be expressed as:
\begin{equation}
\mathbf{K}_t = \underset{\mathbf{K}_t}{\text{argmin}} (\text{rQNSR}(\tilde x_t,x'_t)^2 + \text{MSE}(\tilde x_t,x'_t))
\label{equ:optimization object}
\end{equation}
Eq. \ref{equ:optimization object} transforms the optimization problem into minimizing a function with respect to $\textbf{K}_t$. By taking the derivative of the function with respect to $\textbf{K}_t$ and setting the derivative to zero, we can obtain the analytical solution for $\textbf{K}_t$. The detailed derivation of the formula can be found in supplementary materials.
\begin{equation}
    \mathbf{K}_t = \frac{\sum_{n}^{N}(x'_{t,n}-\mu_t)(\hat x_{t,n}-\mu_t) + \sum_{n}^{N} \frac{\hat x_{t,n}-\mu_t}{x'_{t,n}}}{\sum_{n}^{N}(\hat x_{t,n}-\mu_t)^2 + \sum_{n}^{N}\frac{(\hat x_{t,n}-\mu_t)^2}{x_{t,n}^{'2+}}}
\end{equation}
Where $N$ denotes the number of samples across the optimization
\section{Experiments}
\begin{figure*}
    \centering
    \includegraphics[width=1.\linewidth]
    {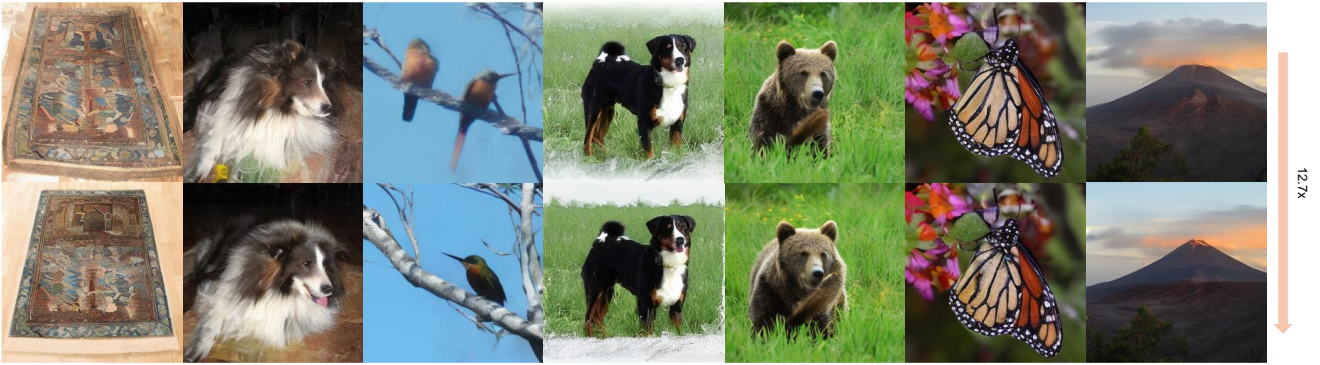}
    \caption{Image generations with our method on DiT. The image sizes are 256 $\times$ 256, with DiT (DDPM, 250 steps, top) and Ours (50 steps, bottom). For more visualizations, please refer to the supplementary materials.}
    \vspace{-3mm}
    \label{fig:visual comparation}
\end{figure*}

\subsection{Experimental Settings}
Our experimental setup closely follows the original configuration used in the Diffusion Transformers (DiTs) study \cite{dit}. We evaluate the performance of our method on the ImageNet dataset \cite{deng2009imagenet} , using pre-trained, class-conditional DiT-XL/2 models \cite{dit} at image resolutions of both 256 $\times$ 256 and 512 $\times$ 512. The DDPM solver \cite{ho2020denoising} with 250 sampling steps is employed for the primary generation process, while additional evaluations with reduced sampling steps of 100 and 50 are conducted to further test the robustness of our approach.

To create a calibration dataset, we generate large-scale samples across the ImageNet classes during the diffusion process, forming a dataset $D_l$. We utilize the TAP algorithm to select a final set for quantization calibration. Specifically, three parallel sampling processes are performed, with each sampling selecting only 1/20 of the samples. This allows us to split $D_l$ into 100 categories, from which we randomly choose 3-10 samples per category, ultimately forming a set of 800 calibration samples—following the implementation of previous works \cite{wu2024ptq4dit}. All experiments are conducted on NVIDIA RTX A100 GPUs, and our code is based on PyTorch \cite{paszke2019pytorch}.

To comprehensively assess the quality of generated images, we employ four evaluation metrics: Fréchet Inception Distance (FID) \cite{heusel2017gans}, spatial FID (sFID) \cite{salimans2016improved,nash2021generating}, Inception Score (IS) \cite{salimans2016improved,barratt2018note}, and Precision. All metrics are computed using the ADM toolkit \cite{dhariwal2021diffusion}. For fair comparison across all methods, including the original models, we sample 10,000 images for ImageNet 256×256 and 5,000 images for ImageNet 512×512, consistent with the standards used in prior studies \cite{nichol2021improved,shang2023post}.

\subsection{Comparison on Performance}
We conduct a comprehensive evaluation of our method against prevalent baselines, being the first to explore the combined effects of quantization and cache. Our benchmarking includes PTQ4DM \cite{shang2023post}, Q-Diffusion \cite{li2023q}, PTQD \cite{he2024ptqd}, Learn-to-Cache \cite{ma2024learning}, RepQ \cite{li2023repq}, and Fora \cite{selvaraju2024fora}. All quantization methods use uniform quantizers, applying channel-wise quantization to weights and tensor-wise quantization to activations, while cache methods store and reuse outputs from self-attention and MLP layers.

Tables \ref{table 1} and \ref{table 2} summarize results for large-scale, class-conditional image generation on ImageNet at resolutions of 256$\times$256 and 512$\times$512. Table \ref{table 1} further demonstrates our method's performance across various timestep settings. Significantly, our findings show that under 8-bit quantization, our method closely aligns with the generative quality of original models, while offering substantial computational efficiency. Figure \ref{fig:comparation} illustrates the efficiency-versus-efficacy trade-off across different configurations, our approach achieves performance comparable to original models (250 timesteps, DDPM) but with markedly reduced computational costs (a 12.7× improvement), presenting a practical solution for high-quality image generation. In all tested settings, our method occupies the upper-left position in the performance-efficiency space, consistently surpassing mainstream alternatives, which reinforces its effectiveness and adaptability.

\begin{table}[]\fontsize{8}{9}\selectfont 
\renewcommand{\tabcolsep}{2pt} 
\caption{Performance comparisonon ImageNet 256 $\times$ 256 with W8A8}
\centering
\label{table 1}
\begin{tabular}{c|c|c|cccc}
\toprule
Steps            & Method         & Speed & FID $\downarrow$  & sFID $\downarrow$  & IS $\uparrow$    &Precision $\uparrow$\\ \hline
\multirow{8}{*}{250} & DDPM              & 1$\times$    & 4.53  & 17.93  & 278.50 & 0.8231                         \\
                     & PTQ4DM         & 2$\times$     & 21.65 & 100.14 & 134.22 & 0.6342                         \\
                     & Q-Diffusion    & 2$\times$     & 5.57  & 18.22  & 227.50 & 0.7612                         \\
                     & PTQD           & 2$\times$     & 5.69  & 18.42  & 224.26 & 0.7594                         \\
                     & RepQ          & 2$\times$     & 4.51  & 18.01  & 264.68 & 0.8076                         \\
                     & PTQ4DiT        & 2$\times$     & 4.63  & 17.72  & 274.86 & 0.8299                         \\
                     & FORA           & 2.06$\times$   & 5.06  & 18.12       & 163.84       & 0.8152                               \\
                     & \cellcolor{gray!20}Q\&C           &\cellcolor{gray!20}  3.12$\times$     &\cellcolor{gray!20}  4.68     &\cellcolor{gray!20}  17.84      &\cellcolor{gray!20} 268.65    &\cellcolor{gray!20}0.8195                               \\ \hline
\multirow{7}{*}{100} & DDPM              & 2.5$\times$    & 5.00  & 17.65  & 274.78 & 0.8068                         \\
                     & PTQ4DM         &  5$\times$     & 15.36 & 79.31  & 172.37 & 0.6926                         \\
                     & Q-Diffusion    &  5$\times$     & 7.93  & 19.46  & 202.84 & 0.7299                         \\
                     & PTQD           & 5$\times$      & 8.12  & 19.64  & 199.00 & 0.7295                         \\
                     & RepQ          & 5$\times$      & 5.20  & 19.87  & 254.70 & 0.7929                         \\
                     & PTQ4DiT        & 5$\times$      & 4.73  & 17.83  & 277.27 & 0.8270                         \\
                     &\cellcolor{gray!20} Q\&C           &\cellcolor{gray!20} 7.23$\times$       &\cellcolor{gray!20} 4.75      &\cellcolor{gray!20}   18.02     &\cellcolor{gray!20} 267.96    &\cellcolor{gray!20} 0.8065                               \\ \hline
\multirow{8}{*}{50}  & DDPM             &      5$\times$ & 5.22  & 17.63  & 237.8 & 0.8056                         \\
                     & PTQ4DM         & 10$\times$      & 17.52 & 84.28  & 154.08 & 0.6574                         \\
                     & Q-Diffusion    & 10$\times$      & 14.61 & 27.57  & 153.01 & 0.6601                         \\
                     & PTQD           & 10$\times$      & 15.21 & 27.52  & 151.60 & 0.6578                         \\
                     & RepQ          & 10$\times$      & 7.17  & 23.67  & 224.83 & 0.7496                         \\
                     & PTQ4DiT        & 10$\times$      & 5.45  & 19.50  & 250.68 & 0.7882                         \\
                     & Learn-to-Cache &   6.3$\times$    & 5.21      &  17.60      &  245.45      & 0.8122                               \\
                     & \cellcolor{gray!20}Q\&C   &\cellcolor{gray!20}  12.7$\times$     &\cellcolor{gray!20}  5.43     &\cellcolor{gray!20} 19.52       & \cellcolor{gray!20}     250.68  &\cellcolor{gray!20} 0.7895                               \\ 
\bottomrule
\end{tabular}
\end{table}

\begin{table}[]\fontsize{8}{9}\selectfont 
\renewcommand{\tabcolsep}{2pt} 
\centering
\caption{Performance on ImageNet 512 $\times$ 512 with W4A8}
\label{table 2}
\begin{tabular}{c|c|c|cccc}
\toprule
Steps            & Method        & Speed & FID $\downarrow$  & sFID $\downarrow$  & IS $\uparrow$    &Precision $\uparrow$\\ \hline
\multirow{8}{*}{100} & DDPM             & 1$\times$    & 9.06  & 37.58 & 239.03 & 0.8300    \\
                     & PTQ4DM        & 2.5$\times$     & 70.63 & 57.73 & 33.82  & 0.4574    \\
                     & Q-Diffusion   & 2.5$\times$     & 62.05 & 57.02 & 29.52  & 0.4786    \\
                     & PTQD          & 2.5 $\times$    & 81.17 & 66.58 & 35.67  & 0.5166    \\
                     & RepQ         & 2.5$\times$     & 62.70 & 73.29 & 31.44  & 0.3606    \\
                     & PTQ4DiT       & 2.5$\times$     & 19.00 & 50.71 & 121.35 & 0.7514    \\
                     
                     & \cellcolor{gray!20}Ours          & \cellcolor{gray!20} 4$\times$      & \cellcolor{gray!20}19.05 & \cellcolor{gray!20}50.71      &  \cellcolor{gray!20}121.11  &\cellcolor{gray!20}0.7533           \\ \hline
\multirow{8}{*}{50}  & DDPM             & 2$\times$      & 11.28 & 41.70 & 213.86 & 0.8100    \\
                     & PTQ4DM        & 5$\times$     & 71.69 & 59.10 & 33.77  & 0.4604    \\
                     & Q-Diffusion   & 5$\times$     & 53.49 & 50.27 & 38.99  & 0.5430    \\
                     & PTQD          & 5$\times$     & 73.45 & 59.14 & 39.63  & 0.5508    \\
                     & RepQ         & 5$\times$     & 65.92 & 74.19 & 30.92  & 0.3542    \\
                     & PTQ4DiT       & 5$\times$     & 19.71 & 52.27 & 118.32 & 0.7336    \\
                     & \cellcolor{gray!20}Q\&C          & \cellcolor{gray!20}6.5$\times$    &\cellcolor{gray!20} 19.71    &\cellcolor{gray!20} 52.26    &\cellcolor{gray!20} 118.45    &\cellcolor{gray!20} 0.7342          \\ \bottomrule
\end{tabular}
\vspace{-3mm}
\end{table}

\subsection{Generality of the Method}
To demonstrate the generality of our method, we also compare it with PTQ4DM \cite{shang2023post} and  APQ-DM \cite{wang2023towards} on LDM across the LSUN-Bedroom and LSUN-Church datasets \cite{yu2015lsun}. The results are as follows.

\begin{table}[h]\fontsize{8}{9}\selectfont 
\renewcommand{\tabcolsep}{2pt} 
\caption{Performance on LDM with W8A8}
\centering
\label{table ldm}
\begin{tabular}{c|c|c|ccc|ccc}
\toprule
\multirow{2}{*}{Step} & \multirow{2}{*}{Method} & \multirow{2}{*}{Speed} & \multicolumn{3}{c|}{LSUN-Bedroom} & \multicolumn{3}{c}{LSUN-Church} \\ \cline{4-9} 
                      &                         &                        & FID $\downarrow$     & sFID  $\downarrow$    & IS  $\uparrow$   & FID $\downarrow$     & sFID $\downarrow$     & IS $\uparrow$    \\ \hline
\multirow{5}{*}{100} & DDIM        & 1$\times$     & 6.39 & 9.45 &2.45&10.98&16.16&2.76      \\
& PTQ4DM        & 2$\times$     & 7.48 & 12.42 &  2.23 & 10.98& 17.28&  2.76     \\
                     & Q-Diffusion   & 2$\times$     & 7.04  &12.24  &  2.27 &12.72& 16.96&   2.72    \\
                     & APQ-DM          &  2$\times$    &  6.46 &  11.82 &  2.55 & 9.04&16.74& 2.84    \\                     
                     & \cellcolor{gray!20}Q\&C          & \cellcolor{gray!20} 3.02 $\times$      & \cellcolor{gray!20}6.52 & \cellcolor{gray!20}11.83      &  \cellcolor{gray!20}2.55         & \cellcolor{gray!20}9.10 & \cellcolor{gray!20}16.73      &  \cellcolor{gray!20} 2.83   \\  \bottomrule
\end{tabular}
\end{table}

\subsection{Visualization of Method Effectiveness}
To examine whether the proposed method effectively improves the sample efficacy of the calibration dataset and mitigates exposure bias, \textbf{we provide comprehensive visualizations in the supplementary materials.} The results clearly demonstrate that TAP and VC significantly enhance each aspect, respectively.

\subsection{Ablation study}
\paragraph{Individual Contributions of TAP and VC}
To assess the effectiveness of TAP and VC, we conducted an ablation study using the W8A8 quantization setup on the ImageNet dataset at a resolution of 256 $\times$ 256, employing 50 sampling timesteps. We evaluated three method variants: (i) Baseline, which leverages the latest quantization and cache techniques, specifically PTQ4DiT \cite{wu2024ptq4dit} combined with Learn-to-Cache \cite{ma2024learning} on DiTs; (ii) Baseline + TAP, which selects an optimized calibration dataset via TAP; and (iii) Baseline + TAP + VC, incorporating both components. The results, presented in Table \ref{table3}, demonstrate performance improvements with each added component, underscoring their effectiveness. 

Notably, the results reveal that TAP and VC contribute significantly to the quality of generated outputs, indicating that our experiments in Section \ref{sect:challenge} accurately identified key challenges in the combined use of quantization and cache, and that our methods effectively address these issues. Specifically, the simple stacking of state-of-the-art quantization and cache methods in the baseline led to a sharp drop in generative quality, whereas adding TAP and VC resulted in substantial improvements, reducing FID by 8.24 and sFID by 6.34, significantly outperforming the baseline.

\begin{table}[]\fontsize{8}{9}\selectfont 
\begin{center}
\caption{ Ablation study on ImageNet 256 $\times$ 256 for 50 timesteps}
\label{table3}
\renewcommand{\tabcolsep}{2pt} 
\begin{tabular}{c|ccccc}
\toprule
Method       & FID $\downarrow$  & sFID $\downarrow$  & IS $\uparrow$    &Precision $\uparrow$ \\ \hline
-             &  5.22  & 17.63  & 237.8 & 0.8056    \\
PTQ4DiT        & 5.45 & 19.50 & 250.68 & 0.7882    \\
Baseline       & 13.67  & 25.86   & 189.65       &   0.7124        \\
+ VC          & 9.65     &22.34       &210.35       &0.7445           \\
+ TAP          & 8.34     & 21.65      &     220.67   & 0.7566          \\
+TAP +VC     &  \cellcolor{gray!20}5.43     &\cellcolor{gray!20} 19.52       &   \cellcolor{gray!20}   250.68  & \cellcolor{gray!20}0.7895                 \\ 
\bottomrule
\end{tabular}
\end{center}
\vspace{-3mm}
\end{table}

\paragraph{Effectiveness of TAP}
To demonstrate the superiority of the TAP method, we compare it with several common clustering methods, covering representative algorithms from partition-based, density-based, and hierarchical clustering approaches. Specifically, we select K-Means \cite{likas2003global}, DBSCAN \cite{deng2020dbscan}, and Agglomerative \cite{murtagh2014ward} Clustering for comparison. The results are as Tab \ref{table:different cluster}.

\begin{table}[h]\fontsize{8}{9}\selectfont 
\begin{center}
\caption{Ablation on TAP with Different Clustering Methods}
\label{table:different cluster}
\renewcommand{\tabcolsep}{4pt} 
\begin{tabular}{c|ccccc}
\toprule
Method       & FID $\downarrow$  & sFID $\downarrow$  & IS $\uparrow$    &Precision $\uparrow$ \\ \hline
Kmeans        &   10.31   &   23.65    & 195.43 & 0.7332     \\
DBSCAN          & 10.12    & 23.21   & 201.35  &0.7365   \\
Agglomerative       & 9.56  &22.13  &202.12 &0.7345         \\
TAP(ours)        &\cellcolor{gray!20} 8.34     &\cellcolor{gray!20} \cellcolor{gray!20}21.65      & \cellcolor{gray!20}    220.67   & \cellcolor{gray!20}0.7566         \\
\bottomrule
\end{tabular}
\end{center}
\vspace{-3mm}
\end{table}

\paragraph{Hyperparameters in TAP}
TAP leverages spatial data distribution and temporal dynamics to construct similarity matrices. To assess the impact of the parameter $\alpha$ in Equ. \ref{equ:similarity matrices} , we conducted ablation experiments, with the results shown in Tab. \ref{table4}

\begin{table}[h]\fontsize{8}{9}\selectfont 
\begin{center}
\caption{ Ablation study on ImageNet 256 $\times$ 256 for similarity Matrices $\alpha$}
\label{table4}
\renewcommand{\tabcolsep}{4pt} 
\begin{tabular}{c|ccccc}
\toprule
$\alpha$       & FID $\downarrow$  & sFID $\downarrow$  & IS $\uparrow$    &Precision $\uparrow$ \\ \hline
0.3         &   5.57   &   19.58    & 248.63 & 0.7823     \\
0.4          & 5.46    & 19.55     & 250.62  &0.7863   \\
0.5       & 5.43  & \textbf{19.52} &\textbf{250.68} &\textbf{0.7895}         \\
0.6        & \textbf{5.36} &  19.56 & 249.53 &0.7875          \\
0.7     & 5.45   & 19.46 & 248.96 &0.7725                 \\ 
\bottomrule
\end{tabular}
\vspace{-5mm}
\end{center}
\end{table}

\section{Related Work}
Enhancing the efficiency of diffusion models has become increasingly necessary due to the high computational cost associated with larger models like DiTs. Quantization and cache mechanisms offer promising approaches to improve the computational efficiency of diffusion models.
\paragraph{Quantization} methods such as Post-Training Quantization (PTQ) have gained attention for their ability to reduce model size and inference time without requiring retraining, making them computationally efficient. Unlike Quantization-Aware Training (QAT), PTQ only requires minimal calibration and can be implemented in a data-free manner by generating calibration datasets using the full-precision model. Techniques like Q-Diffusion \cite{li2023q} apply PTQ methods proposed by BRECQ \cite{li2021brecq} to optimize performance across various datasets, while PTQD \cite{he2024ptqd} mitigates quantization errors by integrating them with diffusion noise. More recent work, such as EfficientDM \cite{he2023efficientdm}, fine-tunes quantized diffusion models using QALoRA \cite{xu2023qa,han2024parameter}, while HQ-DiT \cite{liu2024hq} adopts low-precision floating-point formats, utilizing data distribution analysis and random Hadamard transforms to reduce outliers and enhance quantization performance with minimal computational cost.

\paragraph{Cache} aims to mitigate the computational redundancy in diffusion model inference by leveraging the repetitive nature of sequential diffusion steps. cache in diffusion models leverages the minimal change in high-level features across consecutive steps, enabling reuse of these features while updating only the low-level details. For instance, studies \cite{xu2018deepcache,wimbauer2024cache} reuse feature maps from specific components within U-Net architectures, while \cite{hunter2023fast} focuses on reusing attention maps. Further refinements by \cite{wimbauer2024cache,so2023frdiff,xu2018deepcache} involve adaptive lifetimes for cached features and adjusting scaling to maximize reuse efficiency. Additionally, \cite{zhang2024cross} identifies redundancy in cross-attention during fidelity improvement steps, which can be cached to reduce computation. 

Previous research has accumulated extensive work in both quantization and caching. However, there has been little exploration into how these two acceleration mechanisms can be combined effectively and the challenges that arise from their integration. This work aims to identify these challenges and address them systematically.

\section{Conclusion}
In this paper, we investigated the impact of integrating quantization techniques with cache mechanisms in efficient image generation. Our study identified key challenges when quantization is applied in conjunction with cache strategies, particularly the redundancy in calibration datasets and the exacerbation of exposure bias. To address these challenges, we introduced —Temporal-Aware Parallel Clustering (TAP) for calibration and Variance Compensation (VC) strategy for exposure bias. The results show that the integration of TAP and VC leads to significant improvements in generation quality while maintaining computational efficiency. We believe that our work paves the way for more efficient and effective image generation pipelines. Future research will focus on extending our approach to various types of generative models and further refining the trade-off between computational cost and generation quality.

{
    \small
    \bibliographystyle{ieeenat_fullname}
    \bibliography{main}

\begin{thebibliography}{65}
\providecommand{\natexlab}[1]{#1}
\providecommand{\url}[1]{\texttt{#1}}
\expandafter\ifx\csname urlstyle\endcsname\relax
  \providecommand{\doi}[1]{doi: #1}\else
  \providecommand{\doi}{doi: \begingroup \urlstyle{rm}\Url}\fi

\bibitem[Barratt and Sharma(2018)]{barratt2018note}
Shane Barratt and Rishi Sharma.
\newblock A note on the inception score.
\newblock \emph{arXiv preprint arXiv:1801.01973}, 2018.

\bibitem[Brooks et~al.(2024)Brooks, Peebles, Holmes, DePue, Guo, Jing, Schnurr, Taylor, Luhman, Luhman, Ng, Wang, and Ramesh]{videoworldsimulators2024}
Tim Brooks, Bill Peebles, Connor Holmes, Will DePue, Yufei Guo, Li Jing, David Schnurr, Joe Taylor, Troy Luhman, Eric Luhman, Clarence Ng, Ricky Wang, and Aditya Ramesh.
\newblock Video generation models as world simulators.
\newblock 2024.

\bibitem[Carion et~al.(2020)Carion, Massa, Synnaeve, Usunier, Kirillov, and Zagoruyko]{carion2020end}
Nicolas Carion, Francisco Massa, Gabriel Synnaeve, Nicolas Usunier, Alexander Kirillov, and Sergey Zagoruyko.
\newblock End-to-end object detection with transformers.
\newblock In \emph{European conference on computer vision}, pages 213--229. Springer, 2020.

\bibitem[Chen et~al.(2024)Chen, Shen, Ye, Cao, Tu, Bouganis, Zhao, and Chen]{chen2024delta}
Pengtao Chen, Mingzhu Shen, Peng Ye, Jianjian Cao, Chongjun Tu, Christos-Savvas Bouganis, Yiren Zhao, and Tao Chen.
\newblock Delta-dit: A training-free acceleration method tailored for diffusion transformers.
\newblock \emph{arXiv preprint arXiv:2406.01125}, 2024.

\bibitem[Chen and Cai(2011)]{chen2011large}
Xinlei Chen and Deng Cai.
\newblock Large scale spectral clustering with landmark-based representation.
\newblock In \emph{Proceedings of the AAAI Conference on Artificial Intelligence}, pages 313--318, 2011.

\bibitem[Croitoru et~al.(2023)Croitoru, Hondru, Ionescu, and Shah]{croitoru2023diffusion}
Florinel-Alin Croitoru, Vlad Hondru, Radu~Tudor Ionescu, and Mubarak Shah.
\newblock Diffusion models in vision: A survey.
\newblock \emph{IEEE Transactions on Pattern Analysis and Machine Intelligence}, 45\penalty0 (9):\penalty0 10850--10869, 2023.

\bibitem[Deng(2020)]{deng2020dbscan}
Dingsheng Deng.
\newblock Dbscan clustering algorithm based on density.
\newblock In \emph{2020 7th international forum on electrical engineering and automation (IFEEA)}, pages 949--953. IEEE, 2020.

\bibitem[Deng et~al.(2009)Deng, Dong, Socher, Li, Li, and Fei-Fei]{deng2009imagenet}
Jia Deng, Wei Dong, Richard Socher, Li-Jia Li, Kai Li, and Li Fei-Fei.
\newblock Imagenet: A large-scale hierarchical image database.
\newblock In \emph{2009 IEEE conference on computer vision and pattern recognition}, pages 248--255. Ieee, 2009.

\bibitem[Dhariwal and Nichol(2021)]{dhariwal2021diffusion}
Prafulla Dhariwal and Alexander Nichol.
\newblock Diffusion models beat gans on image synthesis.
\newblock \emph{Advances in neural information processing systems}, 34:\penalty0 8780--8794, 2021.

\bibitem[Feng et~al.(2018)Feng, Yu, and Li]{feng2018faster}
Xu Feng, Wenjian Yu, and Yaohang Li.
\newblock Faster matrix completion using randomized svd.
\newblock In \emph{2018 IEEE 30th International conference on tools with artificial intelligence (ICTAI)}, pages 608--615. IEEE, 2018.

\bibitem[Finkelstein et~al.(2019)Finkelstein, Almog, and Grobman]{finkelstein2019fighting}
Alexander Finkelstein, Uri Almog, and Mark Grobman.
\newblock Fighting quantization bias with bias.
\newblock \emph{arXiv preprint arXiv:1906.03193}, 2019.

\bibitem[Halko et~al.(2011)Halko, Martinsson, and Tropp]{halko2011finding}
Nathan Halko, Per-Gunnar Martinsson, and Joel~A Tropp.
\newblock Finding structure with randomness: Probabilistic algorithms for constructing approximate matrix decompositions.
\newblock \emph{SIAM review}, 53\penalty0 (2):\penalty0 217--288, 2011.

\bibitem[Han et~al.(2024)Han, Gao, Liu, Zhang, et~al.]{han2024parameter}
Zeyu Han, Chao Gao, Jinyang Liu, Sai~Qian Zhang, et~al.
\newblock Parameter-efficient fine-tuning for large models: A comprehensive survey.
\newblock \emph{arXiv preprint arXiv:2403.14608}, 2024.

\bibitem[He et~al.(2023)He, Liu, Wu, Zhou, and Zhuang]{he2023efficientdm}
Yefei He, Jing Liu, Weijia Wu, Hong Zhou, and Bohan Zhuang.
\newblock Efficientdm: Efficient quantization-aware fine-tuning of low-bit diffusion models.
\newblock \emph{arXiv preprint arXiv:2310.03270}, 2023.

\bibitem[He et~al.(2024)He, Liu, Liu, Wu, Zhou, and Zhuang]{he2024ptqd}
Yefei He, Luping Liu, Jing Liu, Weijia Wu, Hong Zhou, and Bohan Zhuang.
\newblock Ptqd: Accurate post-training quantization for diffusion models.
\newblock \emph{Advances in Neural Information Processing Systems}, 36, 2024.

\bibitem[Heusel et~al.(2017)Heusel, Ramsauer, Unterthiner, Nessler, and Hochreiter]{heusel2017gans}
Martin Heusel, Hubert Ramsauer, Thomas Unterthiner, Bernhard Nessler, and Sepp Hochreiter.
\newblock Gans trained by a two time-scale update rule converge to a local nash equilibrium.
\newblock \emph{Advances in neural information processing systems}, 30, 2017.

\bibitem[Ho et~al.(2020)Ho, Jain, and Abbeel]{ho2020denoising}
Jonathan Ho, Ajay Jain, and Pieter Abbeel.
\newblock Denoising diffusion probabilistic models.
\newblock \emph{Advances in neural information processing systems}, 33:\penalty0 6840--6851, 2020.

\bibitem[Hunter et~al.(2023)Hunter, Dudziak, Abdelfattah, Mehrotra, Bhattacharya, and Wen]{hunter2023fast}
Rosco Hunter, {\L}ukasz Dudziak, Mohamed~S Abdelfattah, Abhinav Mehrotra, Sourav Bhattacharya, and Hongkai Wen.
\newblock Fast inference through the reuse of attention maps in diffusion models.
\newblock \emph{arXiv preprint arXiv:2401.01008}, 2023.

\bibitem[Jaiswal et~al.(2023)Jaiswal, Gan, Du, Zhang, Wang, and Yang]{jaiswal2023compressing}
Ajay Jaiswal, Zhe Gan, Xianzhi Du, Bowen Zhang, Zhangyang Wang, and Yinfei Yang.
\newblock Compressing llms: The truth is rarely pure and never simple.
\newblock \emph{arXiv preprint arXiv:2310.01382}, 2023.

\bibitem[Lee et~al.(2023)Lee, Kim, Baek, Hwang, Sung, and Choi]{lee2023enhancing}
Janghwan Lee, Minsoo Kim, Seungcheol Baek, Seok~Joong Hwang, Wonyong Sung, and Jungwook Choi.
\newblock Enhancing computation efficiency in large language models through weight and activation quantization.
\newblock \emph{arXiv preprint arXiv:2311.05161}, 2023.

\bibitem[Li et~al.(2011)Li, Lian, Kwok, and Lu]{li2011time}
Mu Li, Xiao-Chen Lian, James~T Kwok, and Bao-Liang Lu.
\newblock Time and space efficient spectral clustering via column sampling.
\newblock In \emph{CVPR 2011}, pages 2297--2304. IEEE, 2011.

\bibitem[Li et~al.(2023{\natexlab{a}})Li, Qu, Yao, Sun, and Moens]{li2023alleviating}
Mingxiao Li, Tingyu Qu, Ruicong Yao, Wei Sun, and Marie-Francine Moens.
\newblock Alleviating exposure bias in diffusion models through sampling with shifted time steps.
\newblock \emph{arXiv preprint arXiv:2305.15583}, 2023{\natexlab{a}}.

\bibitem[Li et~al.(2023{\natexlab{b}})Li, Liu, Lian, Yang, Dong, Kang, Zhang, and Keutzer]{li2023q}
Xiuyu Li, Yijiang Liu, Long Lian, Huanrui Yang, Zhen Dong, Daniel Kang, Shanghang Zhang, and Kurt Keutzer.
\newblock Q-diffusion: Quantizing diffusion models.
\newblock In \emph{Proceedings of the IEEE/CVF International Conference on Computer Vision}, pages 17535--17545, 2023{\natexlab{b}}.

\bibitem[Li and van~der Schaar(2023)]{li2023error}
Yangming Li and Mihaela van~der Schaar.
\newblock On error propagation of diffusion models.
\newblock In \emph{The Twelfth International Conference on Learning Representations}, 2023.

\bibitem[Li et~al.(2021)Li, Gong, Tan, Yang, Hu, Zhang, Yu, Wang, and Gu]{li2021brecq}
Yuhang Li, Ruihao Gong, Xu Tan, Yang Yang, Peng Hu, Qi Zhang, Fengwei Yu, Wei Wang, and Shi Gu.
\newblock Brecq: Pushing the limit of post-training quantization by block reconstruction.
\newblock \emph{arXiv preprint arXiv:2102.05426}, 2021.

\bibitem[Li et~al.(2024)Li, Xu, Cao, Sun, and Zhang]{li2024q}
Yanjing Li, Sheng Xu, Xianbin Cao, Xiao Sun, and Baochang Zhang.
\newblock Q-dm: An efficient low-bit quantized diffusion model.
\newblock \emph{Advances in Neural Information Processing Systems}, 36, 2024.

\bibitem[Li et~al.(2023{\natexlab{c}})Li, Xiao, Yang, and Gu]{li2023repq}
Zhikai Li, Junrui Xiao, Lianwei Yang, and Qingyi Gu.
\newblock Repq-vit: Scale reparameterization for post-training quantization of vision transformers.
\newblock In \emph{Proceedings of the IEEE/CVF International Conference on Computer Vision}, pages 17227--17236, 2023{\natexlab{c}}.

\bibitem[Likas et~al.(2003)Likas, Vlassis, and Verbeek]{likas2003global}
Aristidis Likas, Nikos Vlassis, and Jakob~J Verbeek.
\newblock The global k-means clustering algorithm.
\newblock \emph{Pattern recognition}, 36\penalty0 (2):\penalty0 451--461, 2003.

\bibitem[Liu et~al.(2023)Liu, Liu, and Cheng]{liu2023oscillation}
Shih-Yang Liu, Zechun Liu, and Kwang-Ting Cheng.
\newblock Oscillation-free quantization for low-bit vision transformers.
\newblock In \emph{International Conference on Machine Learning}, pages 21813--21824. PMLR, 2023.

\bibitem[Liu and Zhang(2024)]{liu2024hq}
Wenxuan Liu and Saiqian Zhang.
\newblock Hq-dit: Efficient diffusion transformer with fp4 hybrid quantization.
\newblock \emph{arXiv preprint arXiv:2405.19751}, 2024.

\bibitem[Liu et~al.(2024)Liu, Li, Xiao, and Gu]{liu2024enhanced}
Xuewen Liu, Zhikai Li, Junrui Xiao, and Qingyi Gu.
\newblock Enhanced distribution alignment for post-training quantization of diffusion models.
\newblock \emph{arXiv preprint arXiv:2401.04585}, 2024.

\bibitem[Lu et~al.(2024)Lu, Zhou, Lin, Liu, Xu, Zhang, Wen, Ren, Gao, Yan, et~al.]{lu2024terdit}
Xudong Lu, Aojun Zhou, Ziyi Lin, Qi Liu, Yuhui Xu, Renrui Zhang, Yafei Wen, Shuai Ren, Peng Gao, Junchi Yan, et~al.
\newblock Terdit: Ternary diffusion models with transformers.
\newblock \emph{arXiv preprint arXiv:2405.14854}, 2024.

\bibitem[Ma et~al.(2024)Ma, Fang, Mi, and Wang]{ma2024learning}
Xinyin Ma, Gongfan Fang, Michael~Bi Mi, and Xinchao Wang.
\newblock Learning-to-cache: Accelerating diffusion transformer via layer caching.
\newblock \emph{arXiv preprint arXiv:2406.01733}, 2024.

\bibitem[Martin et~al.(2018)Martin, Loukas, and Vandergheynst]{martin2018fast}
Lionel Martin, Andreas Loukas, and Pierre Vandergheynst.
\newblock Fast approximate spectral clustering for dynamic networks.
\newblock In \emph{International Conference on Machine Learning}, pages 3423--3432. PMLR, 2018.

\bibitem[Murtagh and Legendre(2014)]{murtagh2014ward}
Fionn Murtagh and Pierre Legendre.
\newblock Ward’s hierarchical agglomerative clustering method: which algorithms implement ward’s criterion?
\newblock \emph{Journal of classification}, 31:\penalty0 274--295, 2014.

\bibitem[Nagel et~al.(2019)Nagel, Baalen, Blankevoort, and Welling]{nagel2019data}
Markus Nagel, Mart~van Baalen, Tijmen Blankevoort, and Max Welling.
\newblock Data-free quantization through weight equalization and bias correction.
\newblock In \emph{Proceedings of the IEEE/CVF International Conference on Computer Vision}, pages 1325--1334, 2019.

\bibitem[Nagel et~al.(2020)Nagel, Amjad, Van~Baalen, Louizos, and Blankevoort]{nagel2020up}
Markus Nagel, Rana~Ali Amjad, Mart Van~Baalen, Christos Louizos, and Tijmen Blankevoort.
\newblock Up or down? adaptive rounding for post-training quantization.
\newblock In \emph{International Conference on Machine Learning}, pages 7197--7206. PMLR, 2020.

\bibitem[Nagel et~al.(2021)Nagel, Fournarakis, Amjad, Bondarenko, Van~Baalen, and Blankevoort]{nagel2021white}
Markus Nagel, Marios Fournarakis, Rana~Ali Amjad, Yelysei Bondarenko, Mart Van~Baalen, and Tijmen Blankevoort.
\newblock A white paper on neural network quantization.
\newblock \emph{arXiv preprint arXiv:2106.08295}, 2021.

\bibitem[Nash et~al.(2021)Nash, Menick, Dieleman, and Battaglia]{nash2021generating}
Charlie Nash, Jacob Menick, Sander Dieleman, and Peter~W Battaglia.
\newblock Generating images with sparse representations.
\newblock \emph{arXiv preprint arXiv:2103.03841}, 2021.

\bibitem[Nichol and Dhariwal(2021)]{nichol2021improved}
Alexander~Quinn Nichol and Prafulla Dhariwal.
\newblock Improved denoising diffusion probabilistic models.
\newblock In \emph{International conference on machine learning}, pages 8162--8171. PMLR, 2021.

\bibitem[Ning et~al.(2023)Ning, Sangineto, Porrello, Calderara, and Cucchiara]{ning2023input}
Mang Ning, Enver Sangineto, Angelo Porrello, Simone Calderara, and Rita Cucchiara.
\newblock Input perturbation reduces exposure bias in diffusion models.
\newblock \emph{arXiv preprint arXiv:2301.11706}, 2023.

\bibitem[Paszke et~al.(2019)Paszke, Gross, Massa, Lerer, Bradbury, Chanan, Killeen, Lin, Gimelshein, Antiga, et~al.]{paszke2019pytorch}
Adam Paszke, Sam Gross, Francisco Massa, Adam Lerer, James Bradbury, Gregory Chanan, Trevor Killeen, Zeming Lin, Natalia Gimelshein, Luca Antiga, et~al.
\newblock Pytorch: An imperative style, high-performance deep learning library.
\newblock \emph{Advances in neural information processing systems}, 32, 2019.

\bibitem[Peebles and Xie(2023)]{dit}
William Peebles and Saining Xie.
\newblock Scalable diffusion models with transformers.
\newblock In \emph{Proceedings of the IEEE/CVF International Conference on Computer Vision}, pages 4195--4205, 2023.

\bibitem[Ranzato et~al.(2015)Ranzato, Chopra, Auli, and Zaremba]{ranzato2015sequence}
Marc'Aurelio Ranzato, Sumit Chopra, Michael Auli, and Wojciech Zaremba.
\newblock Sequence level training with recurrent neural networks.
\newblock \emph{arXiv preprint arXiv:1511.06732}, 2015.

\bibitem[Rennie et~al.(2017)Rennie, Marcheret, Mroueh, Ross, and Goel]{rennie2017self}
Steven~J Rennie, Etienne Marcheret, Youssef Mroueh, Jerret Ross, and Vaibhava Goel.
\newblock Self-critical sequence training for image captioning.
\newblock In \emph{Proceedings of the IEEE conference on computer vision and pattern recognition}, pages 7008--7024, 2017.

\bibitem[Salimans et~al.(2016)Salimans, Goodfellow, Zaremba, Cheung, Radford, and Chen]{salimans2016improved}
Tim Salimans, Ian Goodfellow, Wojciech Zaremba, Vicki Cheung, Alec Radford, and Xi Chen.
\newblock Improved techniques for training gans.
\newblock \emph{Advances in neural information processing systems}, 29, 2016.

\bibitem[Schmidt(2019)]{schmidt2019generalization}
Florian Schmidt.
\newblock Generalization in generation: A closer look at exposure bias.
\newblock \emph{arXiv preprint arXiv:1910.00292}, 2019.

\bibitem[Selvaraju et~al.(2024)Selvaraju, Ding, Chen, Zharkov, and Liang]{selvaraju2024fora}
Pratheba Selvaraju, Tianyu Ding, Tianyi Chen, Ilya Zharkov, and Luming Liang.
\newblock Fora: Fast-forward caching in diffusion transformer acceleration.
\newblock \emph{arXiv preprint arXiv:2407.01425}, 2024.

\bibitem[Shang et~al.(2023)Shang, Yuan, Xie, Wu, and Yan]{shang2023post}
Yuzhang Shang, Zhihang Yuan, Bin Xie, Bingzhe Wu, and Yan Yan.
\newblock Post-training quantization on diffusion models.
\newblock In \emph{Proceedings of the IEEE/CVF conference on computer vision and pattern recognition}, pages 1972--1981, 2023.

\bibitem[So et~al.(2023)So, Lee, and Park]{so2023frdiff}
Junhyuk So, Jungwon Lee, and Eunhyeok Park.
\newblock Frdiff: Feature reuse for universal training-free acceleration of diffusion models.
\newblock \emph{arXiv preprint arXiv:2312.03517}, 2023.

\bibitem[Touvron et~al.(2021)Touvron, Cord, Douze, Massa, Sablayrolles, and J{\'e}gou]{touvron2021training}
Hugo Touvron, Matthieu Cord, Matthijs Douze, Francisco Massa, Alexandre Sablayrolles, and Herv{\'e} J{\'e}gou.
\newblock Training data-efficient image transformers \& distillation through attention.
\newblock In \emph{International conference on machine learning}, pages 10347--10357. PMLR, 2021.

\bibitem[Wang et~al.(2023)Wang, Wang, Xu, Tang, Zhou, and Lu]{wang2023towards}
Changyuan Wang, Ziwei Wang, Xiuwei Xu, Yansong Tang, Jie Zhou, and Jiwen Lu.
\newblock Towards accurate data-free quantization for diffusion models.
\newblock \emph{arXiv preprint arXiv:2305.18723}, 2\penalty0 (5), 2023.

\bibitem[Williams and Aletras(2024)]{williams2024impact}
Miles Williams and Nikolaos Aletras.
\newblock On the impact of calibration data in post-training quantization and pruning.
\newblock In \emph{Proceedings of the 62nd Annual Meeting of the Association for Computational Linguistics (Volume 1: Long Papers)}, pages 10100--10118, 2024.

\bibitem[Wimbauer et~al.(2024)Wimbauer, Wu, Schoenfeld, Dai, Hou, He, Sanakoyeu, Zhang, Tsai, Kohler, et~al.]{wimbauer2024cache}
Felix Wimbauer, Bichen Wu, Edgar Schoenfeld, Xiaoliang Dai, Ji Hou, Zijian He, Artsiom Sanakoyeu, Peizhao Zhang, Sam Tsai, Jonas Kohler, et~al.
\newblock Cache me if you can: Accelerating diffusion models through block caching.
\newblock In \emph{Proceedings of the IEEE/CVF Conference on Computer Vision and Pattern Recognition}, pages 6211--6220, 2024.

\bibitem[Wu et~al.(2024)Wu, Wang, Shang, Shah, and Yan]{wu2024ptq4dit}
Junyi Wu, Haoxuan Wang, Yuzhang Shang, Mubarak Shah, and Yan Yan.
\newblock Ptq4dit: Post-training quantization for diffusion transformers.
\newblock \emph{arXiv preprint arXiv:2405.16005}, 2024.

\bibitem[Wu et~al.(2023)Wu, Xia, Youn, Zheng, Chen, Bakhtiari, Wyatt, Aminabadi, He, Ruwase, et~al.]{wu2023zeroquant}
Xiaoxia Wu, Haojun Xia, Stephen Youn, Zhen Zheng, Shiyang Chen, Arash Bakhtiari, Michael Wyatt, Reza~Yazdani Aminabadi, Yuxiong He, Olatunji Ruwase, et~al.
\newblock Zeroquant (4+ 2): Redefining llms quantization with a new fp6-centric strategy for diverse generative tasks.
\newblock \emph{arXiv preprint arXiv:2312.08583}, 2023.

\bibitem[Xie et~al.(2021)Xie, Wang, Yu, Anandkumar, Alvarez, and Luo]{xie2021segformer}
Enze Xie, Wenhai Wang, Zhiding Yu, Anima Anandkumar, Jose~M Alvarez, and Ping Luo.
\newblock Segformer: Simple and efficient design for semantic segmentation with transformers.
\newblock \emph{Advances in neural information processing systems}, 34:\penalty0 12077--12090, 2021.

\bibitem[Xu et~al.(2018)Xu, Zhu, Liu, Lin, and Liu]{xu2018deepcache}
Mengwei Xu, Mengze Zhu, Yunxin Liu, Felix~Xiaozhu Lin, and Xuanzhe Liu.
\newblock Deepcache: Principled cache for mobile deep vision.
\newblock In \emph{Proceedings of the 24th annual international conference on mobile computing and networking}, pages 129--144, 2018.

\bibitem[Xu et~al.(2023)Xu, Xie, Gu, Chen, Chang, Zhang, Chen, Zhang, and Tian]{xu2023qa}
Yuhui Xu, Lingxi Xie, Xiaotao Gu, Xin Chen, Heng Chang, Hengheng Zhang, Zhengsu Chen, Xiaopeng Zhang, and Qi Tian.
\newblock Qa-lora: Quantization-aware low-rank adaptation of large language models.
\newblock \emph{arXiv preprint arXiv:2309.14717}, 2023.

\bibitem[Yan et~al.(2009)Yan, Huang, and Jordan]{yan2009fast}
Donghui Yan, Ling Huang, and Michael~I Jordan.
\newblock Fast approximate spectral clustering.
\newblock In \emph{Proceedings of the 15th ACM SIGKDD international conference on Knowledge discovery and data mining}, pages 907--916, 2009.

\bibitem[Yang et~al.(2023)Yang, Zhang, Song, Hong, Xu, Zhao, Zhang, Cui, and Yang]{yang2023diffusion}
Ling Yang, Zhilong Zhang, Yang Song, Shenda Hong, Runsheng Xu, Yue Zhao, Wentao Zhang, Bin Cui, and Ming-Hsuan Yang.
\newblock Diffusion models: A comprehensive survey of methods and applications.
\newblock \emph{ACM Computing Surveys}, 56\penalty0 (4):\penalty0 1--39, 2023.

\bibitem[Yu et~al.(2015)Yu, Seff, Zhang, Song, Funkhouser, and Xiao]{yu2015lsun}
Fisher Yu, Ari Seff, Yinda Zhang, Shuran Song, Thomas Funkhouser, and Jianxiong Xiao.
\newblock Lsun: Construction of a large-scale image dataset using deep learning with humans in the loop.
\newblock \emph{arXiv preprint arXiv:1506.03365}, 2015.

\bibitem[Zhang et~al.(2024)Zhang, Liu, Xie, Faccio, Shou, and Schmidhuber]{zhang2024cross}
Wentian Zhang, Haozhe Liu, Jinheng Xie, Francesco Faccio, Mike~Zheng Shou, and J{\"u}rgen Schmidhuber.
\newblock Cross-attention makes inference cumbersome in text-to-image diffusion models.
\newblock \emph{arXiv preprint arXiv:2404.02747}, 2024.

\bibitem[Zhao et~al.(2024{\natexlab{a}})Zhao, Fang, Liu, Rui, Soedarmadji, Li, Lin, Dai, Yan, Yang, et~al.]{zhao2024vidit}
Tianchen Zhao, Tongcheng Fang, Enshu Liu, Wan Rui, Widyadewi Soedarmadji, Shiyao Li, Zinan Lin, Guohao Dai, Shengen Yan, Huazhong Yang, et~al.
\newblock Vidit-q: Efficient and accurate quantization of diffusion transformers for image and video generation.
\newblock \emph{arXiv preprint arXiv:2406.02540}, 2024{\natexlab{a}}.

\bibitem[Zhao et~al.(2024{\natexlab{b}})Zhao, Ning, Fang, Liu, Huang, Lin, Yan, Dai, and Wang]{zhao2024mixdq}
Tianchen Zhao, Xuefei Ning, Tongcheng Fang, Enshu Liu, Guyue Huang, Zinan Lin, Shengen Yan, Guohao Dai, and Yu Wang.
\newblock Mixdq: Memory-efficient few-step text-to-image diffusion models with metric-decoupled mixed precision quantization.
\newblock \emph{arXiv preprint arXiv:2405.17873}, 2024{\natexlab{b}}.

\end{thebibliography}
}

\end{document}